\documentclass{article} 

\PassOptionsToPackage{numbers}{natbib}
\usepackage[main, final]{neurips_2026}   
\makeatletter
\renewcommand{\@noticestring}{AI Agents for Discovery in the Wild (AID-Wild), Workshop at ACM CAIS 2026.}
\makeatother

\usepackage[T1]{fontenc}
\usepackage{lmodern}
\usepackage{courier}
\usepackage{microtype}

\usepackage{multirow}
\usepackage{tabularx}
\usepackage{booktabs}
\usepackage{makecell}

\usepackage{float}
\usepackage{enumitem}
\usepackage{placeins}
\usepackage{subcaption}

\usepackage{tikz}
\usetikzlibrary{arrows.meta,fit,backgrounds,positioning,calc}
\pgfdeclarelayer{background}
\pgfsetlayers{background,main}

\usepackage{listings}
\lstdefinestyle{prompt}{
  basicstyle=\ttfamily\footnotesize,
  breaklines=true,
  columns=flexible,
  keepspaces=true,
}

\usepackage[most]{tcolorbox}
\usepackage{xcolor}
\definecolor{headergray}{RGB}{70,70,70}
\definecolor{bodygray}{RGB}{245,245,245}
\definecolor{bordergray}{RGB}{90,90,90}
\definecolor{darkblue}{rgb}{0, 0, 0.5}
\tcbset{
  promptbox/.style={
    enhanced,
    colback=bodygray,
    colframe=bordergray,
    coltitle=white,
    colbacktitle=headergray,
    fonttitle=\bfseries,
    boxrule=0.8pt,
    arc=2pt,
    left=10pt,
    right=10pt,
    top=8pt,
    bottom=8pt,
    toptitle=4pt,
    bottomtitle=4pt,
    lefttitle=8pt,
    righttitle=8pt,
    breakable
  }
}

\usepackage{setspace}
\usepackage{parskip}

\usepackage{url}
\usepackage{hyperref}
\hypersetup{colorlinks=true, citecolor=darkblue, linkcolor=darkblue, urlcolor=darkblue}

\title{How Do Tool-Augmented LLM Agents Perform on Real-World Energy Analytics Tasks?}

\author{
David Akinpelu\\
Independent Researcher
\And
Akintonde Abbas\thanks{Corresponding author: \texttt{akintonde.abbas@tume.ai}}\\
Tume AI
\And
Rereloluwa Alimi\\
Tume AI
\And
Ayodeji Lana\\
Independent Researcher
}

\begin{document}
\maketitle

\begin{abstract}
While agentic benchmarks have emerged across both general-purpose and domain-specific settings, including finance, coding, law, and drug discovery, energy-domain evaluations remain limited to static knowledge recall. This is a critical gap for a sector that demands live data retrieval, specialized regulatory and market knowledge, and multi-step quantitative reasoning under real-world constraints. Despite its complexity and societal importance, the energy sector remains substantially underserved relative to domains where dynamic, tool-augmented evaluation has matured considerably.

We present an empirical study of tool-augmented LLM agents on real-world energy market analytics tasks. Our evaluation environment consists of 243 expert-curated problems spanning three broad categories: (1) Market Data Retrieval and Analysis, (2) Knowledge Retrieval and Interpretation, and (3) Advanced Quantitative Modeling and Decision Analytics, encompassing tasks such as price and demand analysis, tariff impact modeling, asset revenue and returns estimation, hedging strategy analysis, and optimization modeling, each graded across multiple difficulty levels.

Agents are provided with a configurable suite of domain tools, including live electricity market APIs for major U.S. ISOs, regulatory docket search, utility tariff databases, asset optimization models, and retrieval-augmented generation over energy market documents. To assess the performance of the agents along multiple dimensions, we employ a multi-dimensional evaluation protocol that scores responses on approach correctness, answer accuracy, attribute alignment, and source validity, with category-aware routing to match scoring criteria to question type. We evaluate both closed-source and open-source LLMs, offering a comparative analysis of how model capability and domain tooling interact in a high-stakes professional domain, with key artifacts publicly released.
\end{abstract}

\section{Introduction}
\label{sec:intro}
The energy sector is one of the most analytically demanding domains for AI-assisted decision support. Energy market professionals routinely synthesize heterogeneous, time-sensitive information spanning real-time market prices, complex regulatory frameworks, asset-level financial models, and datasets from ISO/RTO market systems, utility tariffs, interconnection queues, and weather services. Analysts at utilities, independent power producers, regulators, and consulting firms execute these workflows under conditions where errors carry material financial and operational consequences. Large language models (LLMs) have demonstrated strong capabilities in natural language understanding, knowledge retrieval, and structured reasoning~\cite{openai2024gpt4,anthropic2024claude,google2024gemini}. The emergence of tool-augmented agentic frameworks—where models iteratively invoke external tools to retrieve data and execute computations—has further expanded the potential of LLMs in professional analytical workflows~\cite{yao2022react,schick2023toolformer}. Domain-specific benchmarks have begun evaluating such systems in finance, law, software engineering, and drug discovery~\cite{finbench2023,finagentbench2024,guha2023legalbench,jimenez2024swebench,bran2023chemcrow}.

Despite this progress, the energy sector remains largely absent from rigorous agentic evaluation. Most prior AI work in energy focuses on predictive tasks such as load forecasting, renewable generation estimation, and electricity price prediction using supervised learning on historical data~\cite{hong2016probabilistic,weron2014electricity}. WattWorks, a benchmark from the Electric Power Research Institute (EPRI), evaluates LLMs on power system questions but does not assess tool-augmented agents performing multi-step analytical workflows reflective of real analyst practice~\cite{epri2025_llm_power_sector}. Consequently, no benchmark currently evaluates whether LLM agents can execute end-to-end energy analytics workflows under realistic operational constraints. To address this gap, we introduce \textsc{EnergyEvals}, an evaluation framework for tool-augmented LLM agents on real-world energy analytics tasks. The first iteration focuses on U.S. electricity market analytics, with future versions expanding to additional regions and energy sub-domains. This paper makes the following contributions:

\begin{itemize}[leftmargin=*]
\item \textbf{A domain benchmark of 243 expert-curated tasks} spanning three core capability areas -- market data retrieval and analysis, knowledge retrieval and interpretation, and advanced quantitative modeling and decision analytics. Each of these categories contains tasks across three difficulty levels (Easy, Medium, and Hard), which were generated by practitioners with doctoral-level training and combined industry experience exceeding 25 years at leading energy consulting and engineering organizations.

\item \textbf{A configurable agentic evaluation environment} providing agents with nine domain-specific tools, including live ISO/RTO market APIs covering all major U.S. wholesale markets, utility tariff databases, regulatory docket search, renewable energy generation simulation, battery revenue optimization, and retrieval-augmented generation over electricity market reports and market protocol documents.

\item \textbf{A multi-dimensional evaluation protocol} with category-aware rubric routing that assesses approach correctness, answer accuracy, attribute alignment, and source validity through a multiple LLM-as-a-judge framework calibrated to specific quality requirements defined by energy analytics domain experts.

\item \textbf{An empirical study of seven frontier LLMs} spanning closed-source and open-source models, revealing model-specific performance profiles and failure modes that emerge exclusively under realistic agentic task execution in a high-stakes professional domain.

\item \textbf{Public release} of the benchmark dataset, evaluation framework, scoring code, and a subset of agent execution traces, to support reproducibility and community extension.
\end{itemize}

The remainder of this paper is organized as follows. Section~\ref{sec:related} surveys related work. Section~\ref{sec:dataset} describes the benchmark dataset. Section~\ref{sec:agent} presents the agent architecture and tool suite. Section~\ref{sec:evaluation} defines the evaluation protocol. Section~\ref{sec:results} reports experimental results. Conclusions and next steps are covered in Section~\ref{sec:conclusion}.

\section{Related Work}
\label{sec:related}
Research on benchmarks for agentic large language model (LLM) systems has expanded rapidly as tool-use frameworks mature. Early general-purpose benchmarks demonstrate that real-world multi-step reasoning remains difficult even for frontier models. GAIA evaluates web-augmented reasoning tasks where human non-experts solve 92\% of problems while state-of-the-art models score below 30\%~\cite{mialon2023gaia}, and SWE-bench measures software engineering agents solving real GitHub issues~\cite{jimenez2024swebench}. Benchmarks such as AgentBench, ToolBench, $\tau$-bench, and TheAgentCompany consistently reveal large gaps between model reasoning ability and successful task completion across interactive environments, API ecosystems, and simulated enterprise settings~\cite{liu2023agentbench,qin2023toolbench,yao2024taubench,theagentcompany2024}. AstaBench further demonstrates that autonomous scientific discovery remains unsolved~\cite{astabench2024}.

Domain-specific agentic benchmarks have emerged across professional fields, revealing that general capability improvements do not reliably transfer to specialized domains. In finance and enterprise analytics, Finance Agent Benchmark and InvestorBench show that even top models achieve only moderate accuracy~\cite{finagentbench2024,investorbench2025}, while CLASSIC and EnterpriseBench highlight agent struggles with workflow orchestration and tool usage~\cite{topofclass2025,enterprisebench2025}. PaperBench and ScienceAgentBench demonstrate that replicating academic research remains extremely challenging~\cite{paperbench2025,scienceagentbench2025}, and across productivity and specialized domains—including OdysseyBench, ContextBench, MedAgentBench, and LegalAgentBench—performance degrades significantly as tasks require deeper contextual understanding or domain expertise~\cite{odysseybench2025,contextbench2026,medagentbench2025,legalagentbench2025}. SkillsBench covers energy through only three narrow power-system tasks, yet even with curated Skills agents fail more than half of them (47.5\% pass rate) and domain knowledge gaps emerge as a primary failure mode, further motivating purpose-built evaluation frameworks that pair domain-specific tools with tasks representative of real energy analyst workflows~\cite{li2026skillsbench}.

Within the energy domain, most AI research has focused on predictive tasks such as load forecasting, electricity price forecasting, and renewable power generation modeling~\cite{hong2016probabilistic,weron2014electricity,pfenninger2016renewables}. Recent surveys highlight the absence of robust agentic evaluation frameworks for real-world analytical workflows~\cite{amjad2025llmpower}. Some recently published works indicate initial steps to fill this gap. WattWorks from EPRI shows that frontier LLMs perform well on multiple-choice power-sector questions (around 83–86\% accuracy) but decline by roughly 27 percentage points on open-ended technical tasks~\cite{epri2025_llm_power_sector}. ElecBench and smart-grid agent frameworks confirm that retrieval-augmented tools improve but do not fully resolve operational challenges~\cite{elecbench2024,smartgridagents2024}. PFBench presents a benchmark dataset that focuses on power-flow analysis and leverages standard IEEE transmission test cases~\cite{pfbench}. GridMind and GridAgent discuss agents for transmission power flows and contingency analysis and evaluate them on simulation-based test cases~\cite{gridmind,gridagent}. PowerDAG and PowerChain present agentic systems for distribution networks analysis~\cite{powerdag,powerchain}. None of the existing works, however, focus on practical workflows that are part of daily activities for typical energy analysts and decision makers.

Tool-augmented language agents offer a promising paradigm for such workflows. ReAct demonstrated that interleaving reasoning with tool invocation enables iterative solution refinement~\cite{yao2022react}, and Toolformer showed that models can learn autonomous tool invocation~\cite{schick2023toolformer}. Sandboxed code execution and structured tool access further expand agent capabilities~\cite{llminsandbox2026}, though poor tool integration can introduce new reasoning errors~\cite{toolingornottooling2024}. Recent frameworks therefore emphasize multi-dimensional scoring and LLM-as-a-judge methodologies to better diagnose performance~\cite{zheng2023judging,dubois2024alpacafarm,bai2023benchmarking}. Building on these contributions, \textsc{EnergyEvals} evaluates tool-augmented agents on real-world energy analytics tasks involving live market data retrieval, regulatory analysis, and optimization modeling absent from existing benchmarks.

\section{Benchmark Dataset}
\label{sec:dataset}

\subsection{Design Philosophy}

The dataset is designed to evaluate whether tool-augmented LLM agents can execute realistic, end-to-end energy analytics workflows in a professional domain where accuracy, traceability, and quantitative rigor are essential. Rather than testing benchmark-style factual recall, tasks mirror the workflows of practicing energy market analysts, including retrieving live pricing data, interpreting formal regulatory and interconnection documents, and executing multi-step financial models under operationally realistic constraints. Task development was led by domain experts with doctoral-level training and prior professional experience at organizations including McKinsey, ICF, LCG Consulting, and General Electric, representing more than 25 years of combined industry experience. This practitioner grounding is reflected in the prompt design through the use of market-specific terminology (e.g., nodal pricing, ancillary service qualification thresholds, interconnection milestones) and operational constraints such as efficiency parameters, degradation costs, state-of-charge limits, and IRR targets that are typical of client engagements rather than academic exercises. The current release is intentionally scoped to U.S. electricity markets—covering both deregulated and vertically integrated regions—to ensure cross-task comparability while preserving real-world complexity arising from differences in market design, tariff structures, and regulatory and interconnection documentation. Future releases will expand coverage to additional geographies, commodities, and adjacent energy sub-domains.


\subsection{Capability Areas}


The 243-task corpus is organized around three broad capability areas across three difficulty levels (Easy, Medium, Hard), with 107 Data, 86 Knowledge, and 50 Quant.\ tasks respectively (see Appendix~\ref{sec:dataset_breakdown} for the full breakdown).

\begin{enumerate}[leftmargin=*]
  \item \textbf{Market Data Retrieval and Analysis ("Data").} Tasks requiring extraction, aggregation, filtering, and formatting of structured market data from ISO/RTO databases and APIs. Representative analyst functions include day-ahead and real-time price analysis, ancillary service performance evaluation, load and generation dispatch reporting, and cross-market comparisons. Example: \textit{``Show me the monthly average of day-ahead prices for ERCOT Houston hub in 2023 based on your ERCOT database.''}

  \item \textbf{Knowledge Retrieval and Interpretation ("Knowledge").} Tasks requiring navigation of formal regulatory documents, utility tariff filings, market operation manuals, and interconnection procedures to answer precise procedural and structural questions. These tasks test the agent's ability to identify authoritative sources, locate relevant provisions, and interpret regulatory language accurately without fabricating content. Example: \textit{``What are the fees associated with each milestone in the ERCOT generation interconnection process based on the ERCOT fee schedule and Resource Interconnection Handbook?''}

  \item \textbf{Advanced Quantitative Modeling and Decision Analytics ("Quant").} Tasks requiring multi-step analytical reasoning, modeling, and optimization under explicit operational assumptions. Representative functions include battery energy storage revenue estimation, demand-charge impact assessment, internal rate of return (IRR) computation, and optimization-based decision support with explicit constraint specifications. Example: \textit{``If a 4-hour battery earns revenues from arbitrage only in ERCOT West hub over 15 years, what should the \$/MW capex be to earn a 13\% IRR? Assume 81\% roundtrip efficiency, \$25/MWh degradation cost, state-of-charge limits of 10--90\%, and use prices from 2010--2024 as the representative 15-year window.''}
\end{enumerate}



\subsection{Difficulty Stratification}

Tasks are stratified across three difficulty levels to probe increasingly demanding agent behaviors:

\begin{itemize}[leftmargin=*]
  \item \textbf{Easy} - Direct retrieval tasks with explicit source context and limited data transformation. The agent must select and invoke the correct tool but requires minimal multi-step reasoning. Example: \textit{``What detailed fees are associated with each decision point in the NYISO generation interconnection process based on NYISO Manuals 23 and UG21?''}

  \item \textbf{Medium} - Tasks requiring retrieval with or without explicit source hints, combined with moderate aggregation, filtering, or cross-attribute comparison. Example: \textit{``Which PJM price hub had the highest day-ahead average price in January 2024 based on your PJM database?''}

    \item \textbf{Hard} - Tasks requiring multi-step, multi-source reasoning and advanced quantitative modeling under realistic operational assumptions. Agents must correctly sequence tool calls, apply domain-specific constraints, and integrate outputs across multiple reasoning steps. Example: \textit{``If a 4-hour battery earns revenues from arbitrage only in ERCOT West hub over 15 years, what should the \$/MW capex be to earn a 13\% IRR? Assume 81\% roundtrip efficiency, \$25/MWh degradation cost, state-of-charge limits of 10--90\%, and use prices from 2010--2024 as the representative 15-year window.''}
\end{itemize}

\subsection{Paired Prompt Construction}

A central design feature of the dataset is \emph{paired prompt construction}: selected tasks are available in two variants -- one explicitly specifying the information source, and one omitting source specification. This enables controlled evaluation of source-scaffolding effects on agent performance under matched semantic intent. For example, \textit{``What are the participation requirements for regulation service in CAISO based on the latest Business Practice Manual for Market Operations?''} is a task with a specified source. \textit{``What are the participation requirements for regulation service in CAISO?''} is the without-source counterpart.



\section{Agent Architecture and Tool Suite}
\label{sec:agent}
\subsection{ReAct Agent Framework}

Agents are implemented as ReAct-style reasoning-and-acting agents that execute an iterative Thought $\to$ Action $\to$ Observation loop ~\cite{yao2022react}. At each step, the agent produces a natural language reasoning trace (\emph{Thought}), selects and invokes a tool with structured arguments (\emph{Action}), and receives the tool's structured output (\emph{Observation}). The loop terminates when the agent produces a final answer or reaches a configurable maximum iteration budget. This architecture is expressive enough to represent multi-hop retrieval chains, sequential computation pipelines, and iterative refinement strategies without constraining the agent to an execution pattern (see Appendix~\ref{sec:concept_visual} for a conceptual view of the architecture and Appendix~\ref{sec:agent_impl} for implementation details).

\subsection{Model Configurations}

Seven frontier LLMs are evaluated as agent backends (see Appendix~\ref{sec:model_configs} for the full configuration table). Closed-source models (GPT-5.2, GPT-5-mini, Gemini-3.1-Pro, Claude Sonnet~4.6) and open-source models (Kimi-K2.5, Qwen3-Max-Thinking, DeepSeek-V3.2) are all configured with low reasoning effort and temperature~0 for deterministic, reproducible outputs, using off-the-shelf inference APIs without domain adaptation. Reasoning effort is intentionally set to low to evaluate model performance under cost constraints. Future iterations will explore trade-offs between higher reasoning modes and cost implications.

\subsection{Tool Suite}

Agents are given access to a suite of domain-specific tools grouped under nine categories spanning live structured market data (GridStatus API, Database MCP), formal document retrieval (RAG MCP, Dockets, Web Search), domain computation (Battery Optimization, Renewables), and contextual supplementary data (Tariffs, Weather). Tools are registered in a typed registry and exposed as structured JSON Schema function definitions, enabling identical execution across all models. A full tool inventory and descriptions are provided in Appendix~\ref{sec:tool_description}.




All agent executions are traced at the step level as structured JSON artifacts (see Appendix~\ref{sec:observability}), enabling the failure mode analyses in Section~\ref{sec:results}, and a subset is released as a secondary research artifact. Output traces and raw evaluation reports are included in the GitHub repository (https://github.com/Tume-AI/energy-evals)

\section{Evaluation Protocol}
\label{sec:evaluation}

\subsection{Evaluation Dimensions}

Agent responses are assessed across three complementary dimensions that together capture the distinct quality requirements that are typical in the energy analytics domain. Each dimension targets a failure mode that matters in practice but would be invisible under aggregate accuracy-only scoring.

\begin{enumerate}[leftmargin=*]
  \item \textbf{Approach Correctness (1--5).} Does the agent employ an appropriate analytical strategy? This dimension evaluates tool selection, sequencing logic, and whether the agent's reasoning pathway is consistent with how a professional analyst would approach the task. An agent that reaches a correct numerical answer through an inappropriate pathway (for example, by hallucinating data rather than retrieving it from the correct API) receives reduced credit on this dimension.

  \item \textbf{Answer Accuracy / Attribute Alignment (0--1).} Is the final answer factually correct, and does it satisfy all specified constraints, including temporal scope, geographic jurisdiction, entity type, and units of measurement? For tasks that are only quantitative in nature, accuracy is considered based on the difference between the ground truth and the agent's response within an acceptable absolute or relative tolerance. For tasks with a combination of quantitative and qualitative components, a set of up to 5 expected attributes (specific numerical values, named entities, or conclusions) are extracted from the ground truth with an LLM judge and manually reviewed and updated as needed by human domain experts. Three different LLM judges (GPT-5-mini, Gemini-3.1-Flash-Lite and DeepSeek V3.2)   are then used separately to extract attributes from the agent's answer and compare each of the extracted attributes with the expected attributes while considering the defined absolute (e.g. $\epsilon_{\mathrm{abs}}=\pm2$) or relative tolerances (e.g. $\epsilon_{\mathrm{rel}}=\pm10\,\%$) for numerical attributes. The score equals matched\,/\,total attributes, yielding a continuous value in $[0,1]$. This dimension captures failures where an agent retrieves valid data but for the wrong ISO, wrong time period, or wrong entity, as well as simple factual errors.

  \item \textbf{Source Validity (1--5).} Are the data sources cited or implicitly relied upon real, appropriate, and accessible? This dimension penalizes hallucinated source citations, fabricated document version numbers, use of inappropriate or outdated sources, and failures to ground answers in tool-retrieved evidence where the task requires it.
\end{enumerate}

\subsection{Category-Aware Rubric Routing}

Rubric emphasis is adapted to each capability area, reflecting the differential importance of evaluation dimensions across task types. For \emph{Data Retrieval and Analysis} and \emph{Advanced Quantitative Modeling}  tasks, Answer Accuracy is important as a measure of the agent's correctness, since the outcomes of the agent's analysis are expected to closely match the ground truth if all goes well. However, for \emph{Knowledge Retrieval and Interpretation} tasks, Attribute Alignment provides a better measure of correctness because the ground truth will contain multiple attributes that the agent's response is expected to match. Source Validity and Approach Correctness are critical for all the task categories as it is important to verify that the agent is arriving at the answers in a logical way and not via lucky hallucinations.

Rubrics are applied using three different judges, GPT-5-mini, Gemini-3.1-Flash-Lite, and DeepSeek V3.2, with access to human-annotated ground truths and attributes, full agent execution trace, and a structured scoring rubric (see Appendix \ref{sec:prompts}). The median of the three judges' results is taken as the final score for each rubric. Three judges from different providers are considered to avoid bias arising from overreliance on a single judge from a particular provider. The LLM-as-a-judge approach is well-suited to the open-ended, multi-part responses characteristic of professional analytics tasks, which resist reduction to exact-match or template-based scoring ~\cite{zheng2023judging}. Judge outputs include a numeric score on each dimension and a natural language justification, enabling qualitative decision audits.






\subsection{Reported Metrics}

The main reported metrics are as follows.

\begin{itemize}[leftmargin=*]
  \item \textbf{Overall class-balanced means for each dimension with confidence intervals:} $\bar{a}_m$, $\bar{c}_m$, $\bar{v}_m$ aggregated across the 243 tasks considering each category and difficulty level combination. For each model $m$ and question $q$, the judge produces scores on three dimensions: Approach Correctness $a_{m,q} \in [1,5]$, Answer Accuracy $c_{m,q} \in [0,1]$, and Source Validity $v_{m,q} \in [1,5]$. We report class-balanced scores (i.e., weighted average score per category and difficulty level with equal weighting applied to each category-difficulty pair) per dimension, independently without aggregation into a composite score, along with their confidence intervals. Each dimension captures a distinct failure mode, and collapsing them into a single number would obscure the performance profiles that are important to observe. Also, the class-balanced scores account for the different total number of questions in each category. See Appendix \ref{sec:market_data}, \ref{sec:knowledge_retrieval}, and, \ref{sec:quant} for a breakdown by capability areas.
  
  \item \textbf{Efficiency and cost metrics:} Includes total tokens per question, tool calls, and cost per question. These are reported as simple averages. The cost is estimated using input, output, and cached tokens without changes to the default caching behavior of the models. Latency is excluded because network round-trip times vary per provider API and are not a model-capability metric.
  \item \textbf{Task failure rate:} Reflects the percentage of tasks that failed based on four failure model definitions - maxed out iterations, context windows limitations, missing final responses, and clarification requests. This is also reported as a class-balanced metric to avoid placing too much weight on categories with more but easier questions. Also, this failure mode definition focuses on tasks where useful responses were not returned. In subsequent iterations, we will also consider tasks with sub-par responses based on defined thresholds.
\end{itemize}


\section{Results and Analysis}
\label{sec:results}

\subsection{Overall Results}

\begin{table}[htbp]
\centering
\caption{Evaluation Metrics Across Models}
\label{tab:class_balanced_metrics}
\small
\makebox[\textwidth][c]{%
\begin{tabular}{l c c c c @{\hspace{0pt}} c @{\hspace{0pt}} c @{\hspace{0pt}} c}
\toprule
\textbf{Model}
& \textbf{Accuracy}
& \textbf{Approach}
& \begin{tabular}{c}\textbf{Source}\\\textbf{Validity}\end{tabular}
& \textbf{Tokens}
& \begin{tabular}{c}\textbf{Tool}\\\textbf{Calls}\end{tabular}
& \begin{tabular}{c}\textbf{Cost}\\\textbf{Est. (\$)}\end{tabular}
& \begin{tabular}{c}\textbf{Failure}\\\textbf{Rate (\%)}\end{tabular} \\
\midrule
Claude Sonnet 4.6
& $0.56 \pm 0.05$ & $3.94 \pm 0.11$ & $2.72 \pm 0.23$
& 266k & 7.5 & 0.86 & 2.6 \\

Qwen3-Max-Thinking
& $0.44 \pm 0.05$ & $3.74 \pm 0.16$ & $2.24 \pm 0.16$
& 290k & 7.8 & 0.36 & 2.7 \\

DeepSeek V3.2
& $0.43 \pm 0.05$ & $3.42 \pm 0.12$ & $2.16 \pm 0.20$
& 491k & 13.4 & 0.08 & 20.2 \\

Kimi-K2.5
& $0.49 \pm 0.05$ & $3.70 \pm 0.13$ & $2.63 \pm 0.20$
& 453k & 11.4 & 0.07 & 16.5 \\

Gemini-3.1-Pro
& $0.62 \pm 0.07$ & $3.74 \pm 0.16$ & $2.77 \pm 0.30$
& 368k & 6.7 & 0.23 & 3.2 \\

GPT-5-mini
& $0.38 \pm 0.06$ & $3.39 \pm 0.15$ & $3.11 \pm 0.33$
& 101k & 3.7 & 0.01 & 7.5 \\

GPT-5.2
& $0.57 \pm 0.06$ & $3.69 \pm 0.17$ & $4.02 \pm 0.28$
& 191k & 7.3 & 0.12 & 0.8 \\
\bottomrule
\end{tabular}%
}
\end{table}

\begin{figure*}[htbp]
\centering

\begin{subfigure}[t]{0.49\textwidth}
\centering
\includegraphics[width=\linewidth]{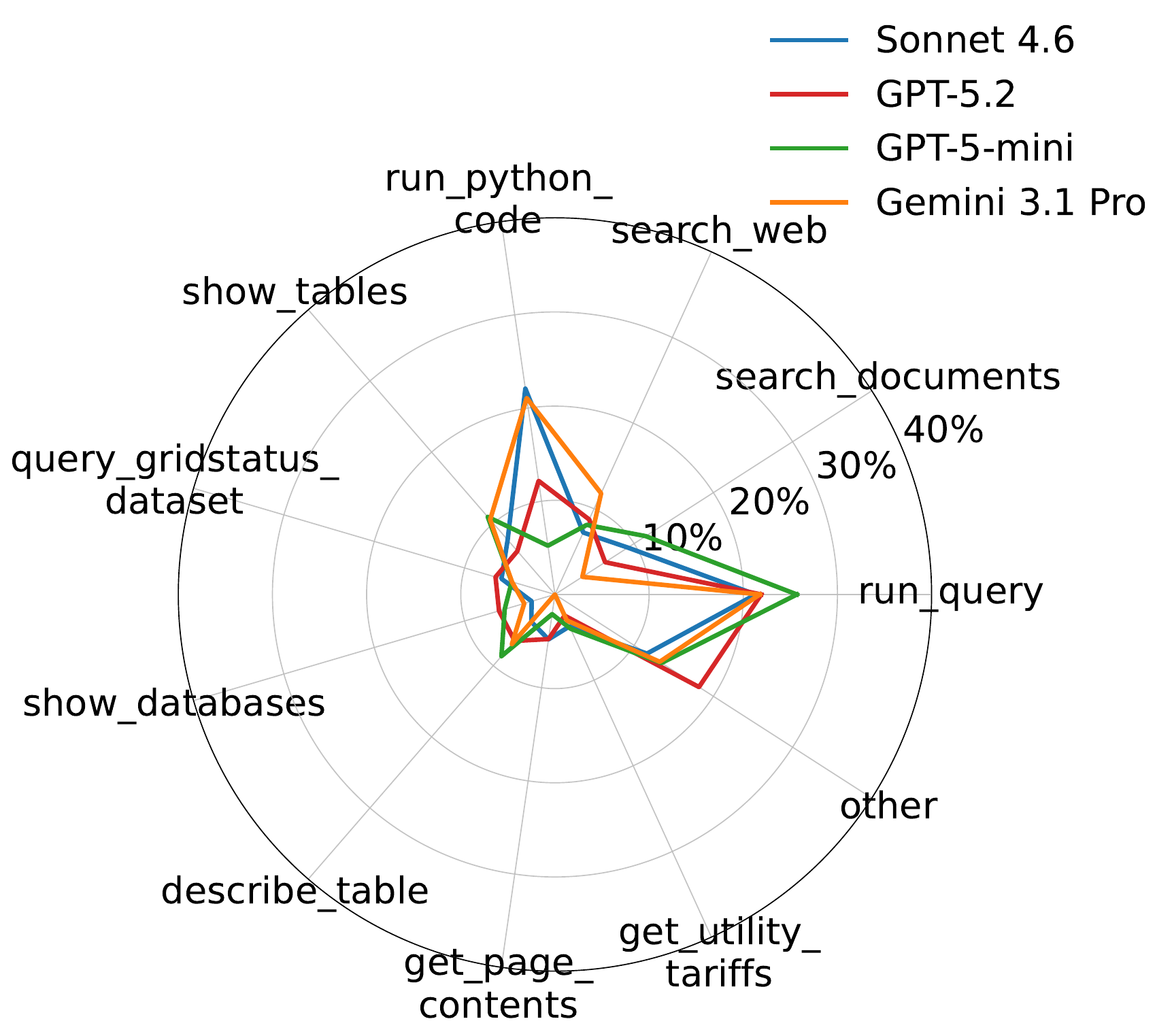}
\caption{Closed-source models}
\label{fig:tool_use_closed}
\end{subfigure}
\hfill
\begin{subfigure}[t]{0.49\textwidth}
\centering
\includegraphics[width=\linewidth]{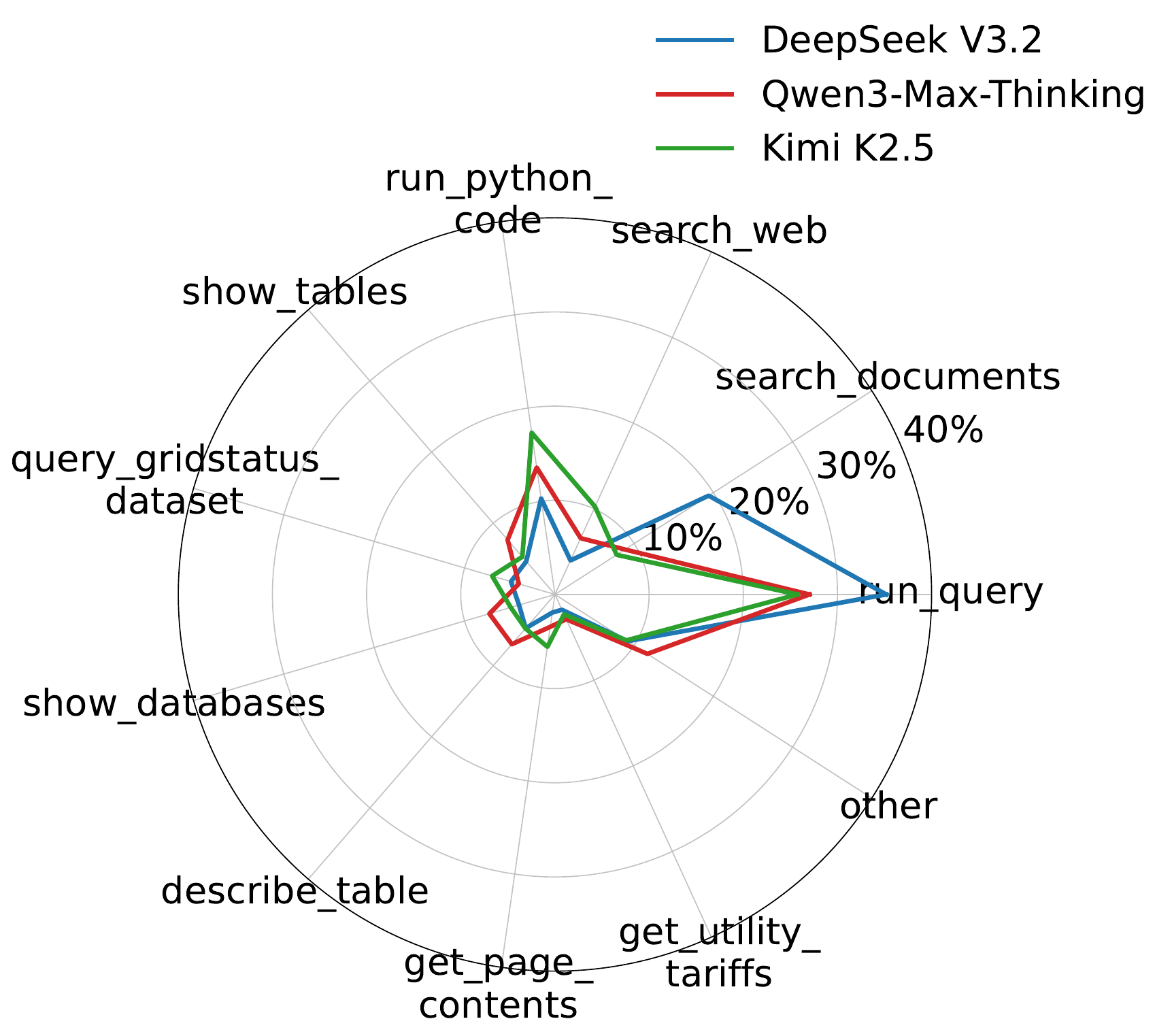}
\caption{Open-source models}
\label{fig:tool_use_open}
\end{subfigure}

\caption{Tool-use distribution across closed-source and open-source models.}
\label{fig:tool_use_chart}
\end{figure*}

Table \ref{tab:class_balanced_metrics} provides a summary of the class-balanced evaluation metrics across all capability areas and difficulty stratifications for the seven frontier models considered. A breakdown of the class-balanced metrics for each capability area is included in the Appendix section (see Appendix \ref{sec:market_data}, \ref{sec:knowledge_retrieval}, and \ref{sec:quant}). 
Figure \ref{fig:tool_use_chart} shows the distribution of tool usage across the different models. The key takeaways from Table \ref{tab:class_balanced_metrics} and Figure \ref{fig:tool_use_chart} are as follows.

\begin{enumerate}[leftmargin=*]
\item \textbf{Closed-source models lead, but overall performance remains unsaturated.} Considering the accuracy results (range: 0 to 1) in Table \ref{tab:class_balanced_metrics}, Gemini-3.1-Pro, GPT-5.2, and Claude Sonnet 4.6 have the best performance values—62\%, 57\%, and 56\%, respectively. The best-performing open-source model is Kimi-K2.5 with a score of 49\%, which is 7 percentage points lower than Claude Sonnet 4.6. However, Kimi-K2.5 achieved this performance at 58\%, 31\%, and 8\% of the costs of GPT-5.2, Gemini-3.1-Pro, and Claude Sonnet 4.6, respectively. The best-performing model, based on accuracy, still shows a 38\% improvement margin, suggesting that domain expertise will continue to play an important role in designing agentic systems with high-accuracy guarantees for energy-domain applications. It is worth noting that Qwen3-Max-Thinking is a cost outlier among open-source models due to its unified reasoning architecture, which defaults to extended chain-of-thought generation, thereby incurring substantially higher token costs than standard open-source models.

\item \textbf{Models exhibit a planning–execution gap in agentic tasks.} The Approach results (range: 0 to 5) from Table \ref{tab:class_balanced_metrics} show that both closed-source and open-source models generally perform reasonably well. GPT-5-mini, which has the lowest performance, has a score equivalent to 68\% (i.e., 3.39/5) compared to a maximum accuracy of 62\%. These results further corroborate the observation that expert guidance remains important in designing agentic systems for energy-domain applications. While both open and closed models are capable of proposing reasonable analytical approaches, reliable execution requires domain-specific knowledge to structure workflows, ensure correct dataset usage, and define appropriate validation criteria.

\item \textbf{Source attribution is not an emergent behavior across current frontier models.} The Source Validity scores (range: 0 to 5) in Table \ref{tab:class_balanced_metrics} are generally low across both closed-source and open-source models. However, OpenAI models tend to exhibit better source attribution behavior. The generally low scores arise because the models do not always include clear source links by default, which could hinder reproducibility of outcomes and limit trust, both of which are particularly important in this context. Explicit prompting with expert guidance will be required to achieve the desired source referencing behavior.

\item \textbf{Context window size has a significant impact on task completion and accuracy.} The models with the highest failure rates are DeepSeek-V3.2, Kimi-K2.5, and GPT-5-mini, as shown in Table \ref{tab:class_balanced_metrics}. The context windows for DeepSeek-V3.2 and Kimi-K2.5 are 163k and 262k tokens, respectively, compared to at least 400k tokens for GPT-5.2, Gemini 3.1 Pro, and Claude Sonnet 4.6 models~\cite{deepinfra_models,openai_models_docs,deepmind_gemini_pro, anthropic_claude_sonnet}. These shorter context windows could be a limitation, especially for knowledge retrieval tasks that require processing significant amounts of information and multi-step problems that require longer contexts to preserve information from each step. Although the Qwen3-Max-Thinking model also has a relatively shorter context window (256k tokens), it interestingly has a considerably smaller failure rate (i.e., 2.7\%). However, its overall accuracy score is significantly lower (i.e., 0.44), implying that tasks that did not fail based on the four failure mode definitions (i.e., maxed-out iterations, context window limitations, missing final answers, and clarification requests) still produced low-accuracy outcomes. GPT-5-mini's failure mode is largely due to excessive requests for additional clarification, which violates the instructions in the system prompt (see Appendix \ref{sec:prompts}).

\item \textbf{Higher token usage is not always correlated with better performance.} From Table \ref{tab:class_balanced_metrics}, some of the least performing models (DeepSeek V3.2 and Kimi K2.5) also had the highest average token usage (494k and 420k, respectively). However, among the top-performing models, Gemini-3.1-Pro has the best performance, but also uses more tokens on average. This suggests that average token usage may not always be a clear indication of superior or inferior performance.

\item \textbf{Tool selection bias appears generally similar between open and closed-source models.} The tool usage distribution charts (Figure \ref{fig:tool_use_chart}) show "run\_query" as the dominant tool across the models. This is because multiple tasks require the retrieval of data from the energy markets database. Excluding the "run\_query" tool, both open and closed source models appear to rely on running Python code to execute tasks. However, DeepSeek V3.2 appears to be more retrieval-heavy.

\end{enumerate}


\subsection{Performance with and without sources specified}
\begin{figure}[h]
\centering
\includegraphics[width=0.6\linewidth]{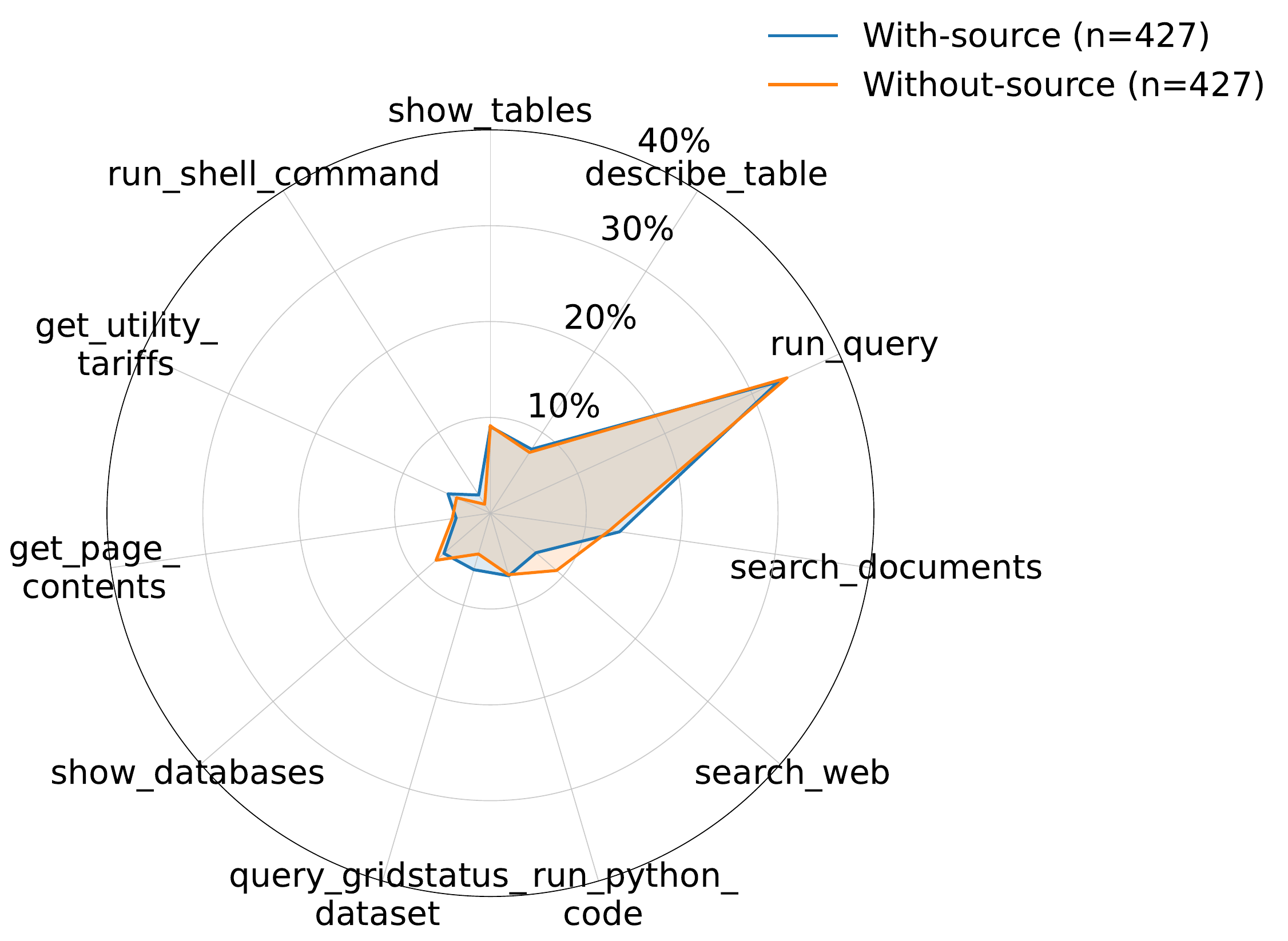}
\caption{Tool use distribution without sources vs with sources specified (n represents 61 questions across 7 models)}
\label{fig:tool_use_with_without_sources}
\end{figure}

As highlighted in the benchmark dataset design philosophy section, paired prompt construction is employed. This provides the basis for measuring the impact of explicitly including sources or tools to use in each question. Table \ref{tab:with_without_sources} shows the class-balanced accuracy and source validity metrics for a subset of 61 tasks (see Appendix \ref{sec:source_task_ids} for the task IDs) that have clear counterparts with and without source. The table presents an interesting observation—the inclusion of sources does not always translate into improved accuracy across all models. This can be explained by a combination of the models' reasoning capabilities and the nature of the questions and tools provided, making it more likely for the models to take similar steps when answering the questions with or without sources specified. Figure \ref{fig:tool_use_with_without_sources} corroborates this, as the distribution of tool usage for both with- and without-source questions is practically the same, except for the uptick in "search\_web" tool usage when sources are not specified. A step-by-step view of the traces across the seven models for a pair of questions (see Appendix \ref{sec:trace_path}) also confirms this. However, source validity performance improves materially for questions with sources, as expected.

\begin{table}[htbp]
\centering
\caption{Evaluation Metrics With vs Without Sources}
\label{tab:with_without_sources}
\small
\setlength{\tabcolsep}{9pt}
\renewcommand{\arraystretch}{1}
\makebox[\textwidth][c]{%
\begin{tabular}{l c c c c}
\toprule
\textbf{Model}
& \multicolumn{2}{c}{\textbf{With Sources}}
& \multicolumn{2}{c}{\textbf{Without Sources}} \\
\cmidrule(lr){2-3} \cmidrule(lr){4-5}
& \textbf{Accuracy}
& \begin{tabular}{c}\textbf{Source}\\\textbf{Validity}\end{tabular}
& \textbf{Accuracy}
& \begin{tabular}{c}\textbf{Source}\\\textbf{Validity}\end{tabular} \\
\midrule
Claude Sonnet 4.6
& $0.68 \pm 0.11$
& $3.58 \pm 0.37$
& $0.72 \pm 0.11$
& $1.92 \pm 0.36$ \\

Qwen3-Max-Thinking
& $0.59 \pm 0.10$
& $2.88 \pm 0.34$
& $0.63 \pm 0.11$
& $1.82 \pm 0.35$ \\

DeepSeek V3.2
& $0.57 \pm 0.11$
& $2.80 \pm 0.33$
& $0.47 \pm 0.09$
& $1.88 \pm 0.39$ \\

Kimi-K2.5
& $0.66 \pm 0.10$
& $3.59 \pm 0.35$
& $0.62 \pm 0.11$
& $2.51 \pm 0.33$ \\

Gemini-3.1-Pro
& $0.69 \pm 0.09$
& $3.88 \pm 0.31$
& $0.71 \pm 0.09$
& $1.84 \pm 0.30$ \\

GPT-5-mini
& $0.47 \pm 0.10$
& $3.42 \pm 0.43$
& $0.49 \pm 0.12$
& $2.23 \pm 0.46$ \\

GPT-5.2
& $0.71 \pm 0.10$
& $4.75 \pm 0.14$
& $0.70 \pm 0.10$
& $3.32 \pm 0.46$ \\
\bottomrule
\end{tabular}%
}
\end{table}

\subsection{Performance with and without selected domain-specific tools}
A subset of 30 tasks (see Appendix \ref{sec:source_task_ids} for the task IDs) was selected from the overall dataset to evaluate performance without domain-specific tools. The 30 tasks represent some of the most challenging in the dataset and are at Medium and Hard difficulty levels across the three capability areas. The Accuracy and Approach metrics are shown in Table \ref{tab:tools_comparison}. The results clearly show that agents perform better when given access to the domain tools across all the models considered. In some cases, the accuracy scores double, emphasizing the importance of domain tools in agentic applications for the energy domain.

\begin{table}[htbp]
\centering
\caption{Evaluation Results With and Without Domain-Specific Tools}
\label{tab:tools_comparison}
\small
{
\setlength{\tabcolsep}{9pt}
\renewcommand{\arraystretch}{1}
\makebox[\textwidth][c]{%
\begin{tabular}{l c c c c}
\toprule
\textbf{Model}
& \multicolumn{2}{c}{\textbf{With Tools}}
& \multicolumn{2}{c}{\textbf{Without Tools}} \\
\cmidrule(lr){2-3} \cmidrule(lr){4-5}
& \textbf{Accuracy}
& \textbf{Approach}
& \textbf{Accuracy}
& \textbf{Approach} \\
\midrule
Claude Sonnet 4.6
& $0.54 \pm 0.16$ & $3.88 \pm 0.23$
& $0.33 \pm 0.13$ & $3.53 \pm 0.26$
\\

Qwen3-Max-Thinking
& $0.28 \pm 0.11$ & $3.60 \pm 0.34$
& $0.11 \pm 0.08$ & $2.78 \pm 0.34$
\\

DeepSeek V3.2
& $0.25 \pm 0.10$ & $3.23 \pm 0.33$
& $0.12 \pm 0.08$ & $2.53 \pm 0.28$
\\

Kimi-K2.5
& $0.28 \pm 0.11$ & $2.85 \pm 0.41$
& $0.14 \pm 0.08$ & $2.98 \pm 0.33$
\\

Gemini-3.1-Pro
& $0.38 \pm 0.15$ & $3.55 \pm 0.31$
& $0.25 \pm 0.12$ & $2.55 \pm 0.34$
\\

GPT-5-mini
& $0.18 \pm 0.10$ & $2.58 \pm 0.36$
& $0.11 \pm 0.08$ & $2.18 \pm 0.28$
\\

GPT-5.2
& $0.48 \pm 0.15$ & $3.53 \pm 0.23$
& $0.25 \pm 0.12$ & $3.00 \pm 0.43$
\\
\bottomrule
\end{tabular}%
}
}
\end{table}

\section{Conclusion}
\label{sec:conclusion}
In previous sections, a rich discussion regarding the performance of tool-augmented LLM agents on real-world energy analytics tasks was presented. The insights show that while tool-augmented agents can tackle some real-world energy analytics tasks, expert guidance is still required to improve the quality of outcomes produced by those agents. This domain expert guidance is what is required to take the performance of these agents from generally good to exceptionally insightful.

We also noted that there are some limitations associated with this first iteration of \textsc{EnergyEvals} and, as such, the following will be captured in future iterations.

1. \textbf{Regional and sub-domain expansion.} The dataset considered in this first iteration of \textsc{EnergyEvals} focuses on the US and tasks relating to electricity markets. While this represents a good starting point, energy analytics is a global phenomenon and goes beyond electricity-related tasks. Subsequent iterations of \textsc{EnergyEvals} will include tasks covering new regions and other energy sub-domains.

2. \textbf{Performance under high reasoning model configurations.} The models considered in this iteration have low reasoning configurations to evaluate performance under resource-constrained scenarios. However, it is possible that other reasoning configuration levels can improve performance. We will investigate this in subsequent iterations. 

\bibliography{references}

\clearpage

\appendix

\section{Appendices}
\renewcommand{\thefigure}{\thesection.\arabic{figure}}
\renewcommand{\thetable}{\thesection.\arabic{table}}
\setcounter{figure}{0}
\setcounter{table}{0}

\subsection{Dataset Breakdown}
\label{sec:dataset_breakdown}

\begin{table}[h]
\centering
\caption{Dataset breakdown by capability area and difficulty level (n=243)}
\label{tab:dataset_breakdown}
\small
\begin{tabular}{lllll}
\toprule
\textbf{Capability Area} & \textbf{Easy} & \textbf{Medium} & \textbf{Hard} & \textbf{Total} \\
\midrule
Data       & 13 & 61 & 33 & 107 \\
Knowledge  & 40 & 43 & 3  & 86  \\
Quant.     & 0  & 8  & 42 & 50  \\
\bottomrule
\end{tabular}
\end{table}

\subsection{Conceptual view of overall evaluation pipeline}
\label{sec:concept_visual}
An illustration of the overall evaluation pipeline is as shown below.
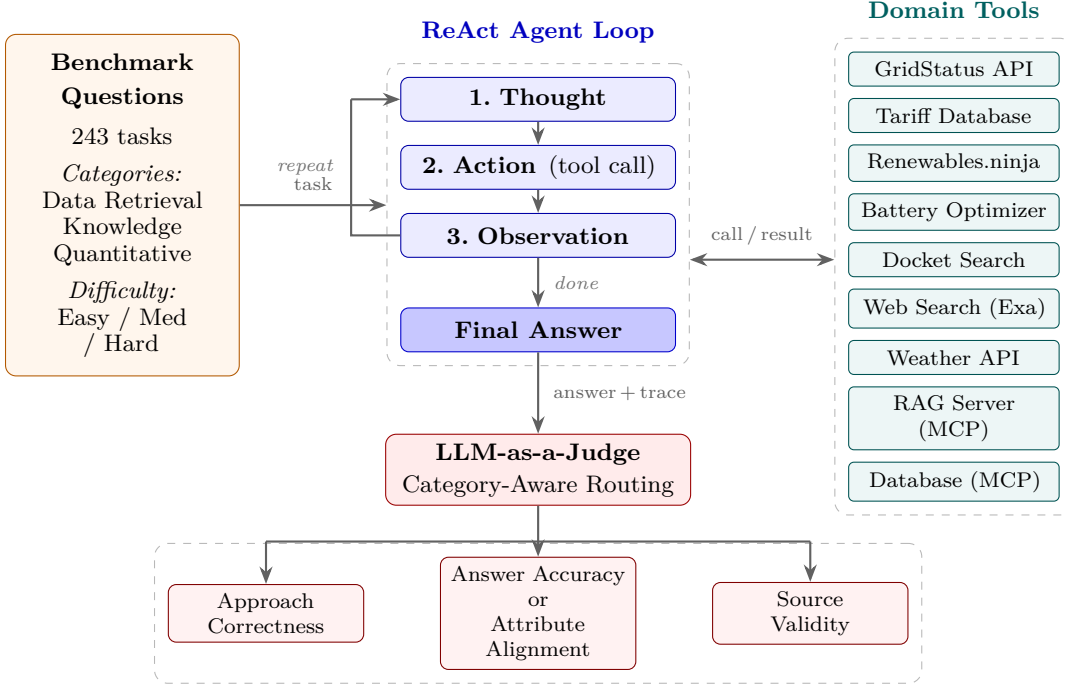
\begin{figure}[H]
\centering
\begin{tikzpicture}[
  font=\small, >=Stealth,
  modelbox/.style={
    rectangle, rounded corners=3pt,
    draw=violet!70!black, fill=violet!8,
    minimum width=1.85cm, minimum height=0.7cm,
    align=center, font=\scriptsize},
  stepbox/.style={
    rectangle, rounded corners=3pt,
    draw=blue!65!black, fill=blue!8,
    text width=3.4cm, minimum height=0.58cm,
    align=center},
  answerbox/.style={
    rectangle, rounded corners=3pt,
    draw=blue!80!black, fill=blue!22,
    text width=3.4cm, minimum height=0.58cm,
    align=center, font=\small\bfseries},
  toolbox/.style={
    rectangle, rounded corners=2pt,
    draw=teal!65!black, fill=teal!7,
    text width=2.55cm, minimum height=0.45cm,
    align=center, font=\footnotesize},
  databox/.style={
    rectangle, rounded corners=4pt,
    draw=orange!70!black, fill=orange!7,
    text width=2.6cm, align=center,
    inner sep=7pt},
  judgebox/.style={
    rectangle, rounded corners=4pt,
    draw=red!60!black, fill=red!8,
    text width=3.8cm, minimum height=0.72cm,
    align=center},
  evalbox/.style={
    rectangle, rounded corners=3pt,
    draw=red!50!black, fill=red!5,
    text width=2.35cm, minimum height=0.72cm,
    align=center, font=\footnotesize},
  grpbg/.style={
    rectangle, rounded corners=5pt,
    draw=black!28, dashed, fill=none,
    inner sep=5pt},
  arr/.style={->, thick, black!62},
  biarr/.style={<->, thick, black!62},
]

\node[stepbox]   (thought) at (0, -2.6)  {\textbf{1.~Thought}};
\node[stepbox]   (action)  at (0, -3.5)  {\textbf{2.~Action}\enspace{\small(tool call)}};
\node[stepbox]   (obs)     at (0, -4.4)  {\textbf{3.~Observation}};
\node[answerbox] (ans)     at (0, -5.65) {Final Answer};

\draw[arr] (thought) -- (action);
\draw[arr] (action)  -- (obs);
\draw[arr] (obs) -- node[right=2pt, font=\scriptsize, text=black!55]
  {\textit{done}} (ans);

\draw[thick, black!62]
  (obs.west) -- ++(-0.65, 0) coordinate (loopleft)
             -- ++(0, 1.8)   coordinate (looptop);
\draw[arr] (looptop) -- (thought.west);
\node[font=\scriptsize, text=black!55, left=3pt]
  at ($(loopleft)!0.5!(looptop)$) {\textit{repeat}};

\begin{pgfonlayer}{background}
  \node[grpbg, fit=(thought)(action)(obs)(ans)] (reactgrp) {};
\end{pgfonlayer}
\node[above=4pt of reactgrp.north, font=\small\bfseries, text=blue!75!black]
  {ReAct Agent Loop};

\node[databox] (dataset) at (-5.5, -4.0) {
  \textbf{Benchmark}\\[3pt]\textbf{Questions}\\[5pt]
  243 tasks\\[4pt]
  \emph{Categories:}\\
  Data Retrieval\\
  Knowledge\\
  Quantitative\\[4pt]
  \emph{Difficulty:}\\
  Easy / Med / Hard
};

\draw[arr] (dataset.east) --
  node[above=2pt, font=\scriptsize] {task}
  ([yshift=0pt]reactgrp.west |- dataset.east);

\node[toolbox] (t1) at (5.5, -2.2) {GridStatus API};
\node[toolbox, below=0.15cm of t1] (t2) {Tariff Database};
\node[toolbox, below=0.15cm of t2] (t3) {Renewables.ninja};
\node[toolbox, below=0.15cm of t3] (t4) {Battery Optimizer};
\node[toolbox, below=0.15cm of t4] (t5) {Docket Search};
\node[toolbox, below=0.15cm of t5] (t6) {Web Search (Exa)};
\node[toolbox, below=0.15cm of t6] (t7) {Weather API};
\node[toolbox, below=0.15cm of t7] (t8) {RAG Server (MCP)};
\node[toolbox, below=0.15cm of t8] (t9) {Database (MCP)};

\begin{pgfonlayer}{background}
  \node[grpbg, fit=(t1)(t2)(t3)(t4)(t5)(t6)(t7)(t8)(t9)] (toolgrp) {};
\end{pgfonlayer}
\node[above=4pt of toolgrp.north, font=\small\bfseries, text=teal!75!black]
  {Domain Tools};

\draw[biarr]
  ([yshift=0pt]reactgrp.east |- t5.west) --
  node[above=2pt, font=\scriptsize] {call\,/\,result}
  (toolgrp.west |- t5.west);  

\node[judgebox] (judge) at (0, -7.5) {
  \textbf{LLM-as-a-Judge}\\[2pt]
  {\small Category-Aware Routing}
};

\draw[arr] (ans.south) --
  node[right=2pt, font=\scriptsize] {answer\,+\,trace}
  (judge.north);

\node[evalbox] (e1) at (-3.6, -9.4) {Approach\\Correctness};

\node[evalbox] (e2) at (0, -9.4) {
  Answer Accuracy\\
  or\\
  Attribute Alignment
};

\node[evalbox] (e3) at (3.6, -9.4) {Source\\Validity};

\begin{pgfonlayer}{background}
  \node[grpbg, fit=(e1)(e2)(e3)] (evalgrp) {};
\end{pgfonlayer}

\coordinate (busmid) at (0, -8.45);

\draw[thick, black!62] (judge.south) -- (busmid);

\draw[thick, black!62]
  ([xshift=-3.6cm]busmid) --
  ([xshift= 3.6cm]busmid);

\draw[arr]
  ([xshift=-3.6cm]busmid) -- (e1.north);

\draw[arr]
  (busmid) -- (e2.north);

\draw[arr]
  ([xshift=3.6cm]busmid) -- (e3.north);

\end{tikzpicture}
\caption{Conceptual overview of the evaluation pipeline. A question from the benchmark dataset is presented to a ReAct agent backed by one of seven LLMs. The agent iterates through \textbf{Thought}, \textbf{Action} (tool call), and \textbf{Observation} steps, invoking domain tools as needed, until it emits a Final Answer. The answer and execution trace are then scored by the three LLM judges across four dimensions using category-aware rubric routing.}
\label{fig:architecture}
\end{figure}

\clearpage

\subsection{Model Configurations}
\label{sec:model_configs}

\begin{table}[h]
\centering
\caption{Models evaluated and inference configurations}
\label{tab:models}
\small
\begin{tabular}{lllll}
\toprule
\textbf{Model} & \textbf{Provider} & \textbf{Open Source} & \textbf{Reasoning} & \textbf{Reasoning Level} \\
\midrule
GPT-5.2       & OpenAI & No   & Yes & Low \\
GPT-5-mini    & OpenAI & No   & Yes & Low \\
Gemini-3.1-Pro    & Google & No   & Yes & Low \\
Claude Sonnet 4.6 & Anthropic & No & Yes   & Low   \\
Kimi-K2.5     & Moonshot & Yes  & Yes      & N/A   \\
Qwen3-Max-Thinking & Alibaba & Yes & Yes   & N/A   \\
DeepSeek-V3.2 & DeepSeek  & Yes & Yes      & N/A   \\
\bottomrule
\end{tabular}
\end{table}

GPT, Gemini, and Sonnet models are configured with low reasoning effort to evaluate performance under compute-efficient inference conditions. In subsequent releases, high reasoning level configurations will be examined. All models are evaluated with temperature set as 0 for deterministic, reproducible outputs. No system-level fine-tuning or domain adaptation is applied; all models are used off-the-shelf via their respective inference APIs (with DeepInfra APIs used for all the open source models~\cite{deepinfra_models}).

\subsection{Agent Implementation Details}
\label{sec:agent_impl}

The ReAct agent is implemented through a provider abstraction layer that wraps model-specific API differences including function-calling schemas, tool-use message blocks, response parsing logic, and streaming behavior behind a common interface. This design enables identical benchmark execution across all evaluated models without modification to the agent loop or tool suite.

\subsection{Tool Suite Overview}
\label{sec:tool_suite_overview}

\begin{table}[h]
\centering
\caption{Tool suite available to agents}
\label{tab:tools}
\small
\begin{tabular}{lll}
\toprule
\textbf{Tool Category} & \textbf{Data Source} & \textbf{Coverage} \\
\midrule
GridStatus API  & GridStatus.io      & All US wholesale electricity markets \\
Tariffs         & OpenEI Tariffs API    & U.S. utility tariffs \\
Renewables      & Renewables.ninja   & Solar/wind generation simulation \\
Battery Optim.  & N/A      & Arbitrage-only revenues for battery projects \\
Dockets         & FERC; state PUCs & Federal and 7 other state jurisdictions \\
Web Search      & Exa API            & Open web \\
Weather         & OpenWeatherMap     & Current \& forecast \\
RAG (MCP)       & Document corpus    & Market reports and manuals\\
Database (MCP)  & Market data portals      & ERCOT, NYISO, PJM and ISONE markets \\
\bottomrule
\end{tabular}
\end{table}

\subsection{Observability and Trace Collection}
\label{sec:observability}

All agent executions are traced at the step level, capturing the complete sequence of thought, action, observation, and answer events as structured JSON artifacts. Traces record tool call arguments and raw responses alongside token-level timing and iteration counts. These traces serve two functions: they enable the failure mode discussions in Section~\ref{sec:results}, and they constitute a secondary research artifact released alongside the benchmark.

\clearpage

\subsection{Tool description}
\label{sec:tool_description}
\begin{table}[H]
\centering
\caption{Tool inventory grouped by category}
\label{tab:tool_inventory_grouped}
\small
\begin{tabularx}{\textwidth}{p{0.18\textwidth} p{0.30\textwidth} X}
\toprule
\textbf{Tool category} & \textbf{Tool} & \textbf{Description} \\
\midrule

  \multirow{4}{*}{System}
  & list\_files & Lists files/directories in a specified path, optionally
  recursively. \\
  & grep\_files & Searches files for text patterns with optional glob/
  path filters. \\
  & run\_python\_code & Executes Python code in a sandboxed environment
  and returns output/errors. \\
  & run\_shell\_command & Executes shell commands in a controlled
  environment and returns stdout/stderr. \\

\multirow{3}{*}{GridStatus API}
& list\_gridstatus\_datasets & Lists available GridStatus datasets (ID, name, description). \\
& inspect\_gridstatus\_dataset & Returns schema/metadata for a specific GridStatus dataset. \\
& query\_gridstatus\_dataset & Queries a GridStatus dataset with filters/time bounds and returns results. \\

\multirow{1}{*}{Tariffs}
& get\_utility\_tariffs & Retrieves utility tariff/rate records from OpenEI IURDB. \\

\multirow{2}{*}{Renewables}
& get\_solar\_profile & Returns hourly solar generation profile (capacity factors) for a location/date range. \\
& get\_wind\_profile & Returns hourly wind generation profile (capacity factors) for a location/date range. \\

\multirow{1}{*}{Battery Optim.}
& battery\_revenue\_optimization & Solves battery dispatch/arbitrage optimization and outputs revenue metrics/profile. \\

\multirow{8}{*}{Dockets}
& search\_ferc\_dockets & Searches FERC dockets/filings. \\
& search\_dc\_dockets & Searches District of Columbia PSC dockets. \\
& search\_maryland\_dockets & Searches Maryland PSC dockets. \\
& search\_new\_york\_dockets & Searches New York PSC dockets. \\
& search\_north\_carolina\_dockets & Searches North Carolina Utilities Commission dockets. \\
& search\_south\_carolina\_dockets & Searches South Carolina PSC dockets. \\
& search\_texas\_dockets & Searches Texas PUCT dockets/filings. \\
& search\_virginia\_dockets & Searches Virginia SCC dockets. \\

\multirow{2}{*}{Web Search}
& search\_web & Runs web search over external sources. \\
& get\_page\_contents & Fetches and extracts content from specified URLs. \\

\multirow{5}{*}{Weather}
& geocode\_location & Converts location names to latitude/longitude. \\
& get\_current\_weather & Returns current weather conditions for a location. \\
& get\_forecast & Returns short-term weather forecast for a location. \\
& get\_historical\_weather & Returns historical weather over a specified period. \\
& get\_air\_pollution & Returns air-quality and pollutant metrics for a location. \\

\multirow{1}{*}{RAG (MCP)}
& search\_documents & Retrieves relevant passages from indexed document corpora (MCP RAG). \\

\multirow{7}{*}{Database (MCP)}
& show\_databases & Lists accessible databases. \\
& show\_tables & Lists tables in a selected database/schema. \\
& describe\_table & Returns table schema/column metadata. \\
& show\_indexes & Shows table indexes for query planning. \\
& run\_query & Executes SQL query against connected database. \\
& inspect\_query & Provides query inspection/validation metadata. \\
& preview\_table & Returns a row preview/sample from a table. \\

\bottomrule
\end{tabularx}
\end{table}

\clearpage

\subsection{System and evaluation prompts}
\label{sec:prompts}
Here are the different prompts used for the benchmark and evaluation runs. They are also included in the publicly available repository.

\begin{tcolorbox}[promptbox,title=Agent\ System\ Prompt]
\ttfamily
\small
You are an Expert Energy Analyst.

Use your best effort to answer each question with only one attempt.

No room for back and forths with the user
\end{tcolorbox}
\vspace{1em}

\begin{tcolorbox}[promptbox,title=Judge\ System\ Prompt]
\ttfamily
\small
You are a strict evaluator of answers relating to energy markets analysis.

Follow expert industry standards.

Your output MUST exactly match the provided output schema.

Do not add extra fields or surrounding text.
\end{tcolorbox}

\vspace{1em}

\begin{tcolorbox}[promptbox,title=Approach Evaluation Prompt]
\ttfamily
\small
You are evaluating the approach correctness of how an AI agent obtained answers to an energy market related question
and not the correctness of the answer itself.

In addition to question, you also have a summary of the suggested approach provided by an expert and a trace
of the steps the agent took to answer the question which you can use to infer the agent's approach to answering
the question

Question: \\
\{question\}

Suggested Approach (Ground Truth): \\
\{suggested\_steps\}

Agent's Steps: \\
\{agent\_steps\_trace\}

Evaluate:
\begin{itemize}
    \item Correct problem framing
    \item Appropriate data sources (ISO postings, tariffs, settlement data, APIs)
    \item Logical analytical steps
    \item Correct tool usage (if applicable)
\end{itemize}

Rating scale: \\
5=expert-like, 4=minor issues, 3=notable gaps, 2=major flaws, 1=wrong approach
\end{tcolorbox}

\vspace{1em}

\begin{tcolorbox}[promptbox,title=Accuracy Evaluation Prompt]
\ttfamily
\small
You are evaluating the factual and numerical accuracy of an AI agent's answer to a question relating to energy markets
analysis.

Question: \\
\{question\}

Expected Answer (Ground Truth): \\
\{expected\_answer\}

Agent's Answer: \\
\{agent\_answer\}

Evaluate:
\begin{itemize}
    \item Numerical correctness (values, sign, magnitude, units, time basis)
    \item Factual alignment (market/ISO, node/zone, product, settlement type etc.)
    \item Completeness of key facts
\end{itemize}

Tolerance: \\
Allow $\leq$ \{abs\_tol\} absolute error OR $\leq$ \{rel\_tol\}\% relative error unless exactness is required.
\end{tcolorbox}

\vspace{1em}

\begin{tcolorbox}[promptbox,title=Source Evaluation Prompt]
\ttfamily
\small
You are evaluating the following two things only.

1. Explicit inclusion of sources in an AI agent's answer to a question relating
to energy markets analysis.

2. Relevance of the included sources for the question

You can extract or infer relevant sources from the question itself or from the suggested approach ground truth

Do not penalize for not explicitly adding queries or code for pulling data for verification as long as the source
specified is consistent with what the agent has access to and is plausible

Internal databases are based on data from authoritative external sources and as such, the internal databases are
equivalent to external authoritative sources (e.g. market portals) and should be treated as such

Question: \\
\{question\}

Suggested Steps: \\
\{suggested\_steps\}

Agent's Answer: \\
\{agent\_answer\}

Evaluate:
\begin{itemize}
    \item Authority of sources
    \item Alignment with expected sources
    \item Appropriateness for the claim
    \item Missing citations when required
\end{itemize}
\end{tcolorbox}

\vspace{1em}

\begin{tcolorbox}[promptbox,title=Attribue Evaluation Prompt]
\ttfamily
\small
You are evaluating attribute alignment of an AI agent's answer against a canonical set of expected attributes.

Question: \\
\{question\}

Expected Attributes (canonical, JSON): \\
\{expected\_attributes\_json\}

Agent's Answer: \\
\{agent\_answer\}

For each expected attribute, decide whether the agent answer contains the correct value
or a reasonable equivalent, respecting units and time basis.

Tolerance: \\
For numeric attributes, allow $\leq$ \{abs\_tol\} absolute error OR $\leq$ \{rel\_tol\}\% relative error unless exactness is required.
\end{tcolorbox}

\vspace{1em}

\begin{tcolorbox}[promptbox,title=Attribute Extraction Prompt]
\ttfamily
\small
You are generating a canonical attribute set for evaluating an AI agent answer to an energy market question.

Extract no more than 5 high-value attributes from the expected answer.

Each attribute should be specific, evaluable, and tied to the question intent.

Prefer attributes that are most critical to correctness.

Question: \\
\{question\}

Expected Answer: \\
\{expected\_answer\}
\end{tcolorbox}

\subsection{Results for Market Data Retrieval and Analysis Tasks}
\label{sec:market_data}
Compared with the overall results in Table~\ref{tab:class_balanced_metrics}, accuracy scores are higher for market data retrieval and analysis tasks. This suggests that the models generally perform better on this category of questions. Failure rates are also lower for most models, with the exception of GPT-5-mini, whose failures appear to be driven largely by clarification requests rather than by errors specific to this task category.

\begin{table}[!htbp]
\centering
\caption{Evaluation Metrics for Market Data Retrieval and Analysis Tasks}
\label{tab:data_class_balanced_metrics}
\small
{
\setlength{\tabcolsep}{5pt}
\renewcommand{\arraystretch}{1.15}
\makebox[\textwidth][c]{%
\begin{tabular}{l c c c c @{\hspace{3pt}} c @{\hspace{3pt}} c @{\hspace{3pt}} c}
\toprule
\textbf{Model}
& \textbf{Accuracy}
& \textbf{Approach}
& \begin{tabular}{c}\textbf{Source}\\\textbf{Validity}\end{tabular}
& \textbf{Tokens}
& \begin{tabular}{c}\textbf{Tool}\\\textbf{Calls}\end{tabular}
& \begin{tabular}{c}\textbf{Cost}\\\textbf{Est. (\$)}\end{tabular}
& \begin{tabular}{c}\textbf{Failure}\\\textbf{Rate (\%)}\end{tabular} \\
\midrule
Claude Sonnet 4.6
& $0.73 \pm 0.06$ & $3.99 \pm 0.15$ & $2.44 \pm 0.33$
& 261k & 7.2 & 0.81 & 1.2 \\

Qwen3-Max-Thinking
& $0.67 \pm 0.07$ & $4.18 \pm 0.17$ & $2.39 \pm 0.28$
& 293k & 8.5 & 0.36 & 1.9 \\

DeepSeek V3.2
& $0.61 \pm 0.06$ & $3.93 \pm 0.17$ & $2.14 \pm 0.26$
& 560k & 14.9 & 0.09 & 6.8 \\

Kimi-K2.5
& $0.66 \pm 0.07$ & $3.94 \pm 0.18$ & $2.73 \pm 0.34$
& 494k & 11.9 & 0.07 & 6.9 \\

Gemini-3.1-Pro
& $0.77 \pm 0.05$ & $4.04 \pm 0.14$ & $2.83 \pm 0.31$
& 273k & 6.6 & 0.17 & 1.2 \\

GPT-5-mini
& $0.53 \pm 0.09$ & $3.86 \pm 0.19$ & $3.00 \pm 0.33$
& 108k & 4.6 & 0.01 & 4.9 \\

GPT-5.2
& $0.75 \pm 0.05$ & $4.08 \pm 0.17$ & $4.22 \pm 0.21$
& 264k & 9.8 & 0.13 & 0 \\
\bottomrule
\end{tabular}%
}
}
\end{table}

\FloatBarrier

\subsection{Results for Knowledge Retrieval and Interpretation Tasks}
\label{sec:knowledge_retrieval}

Scores in Table~\ref{tab:knowledge_class_balanced_metrics} are generally lower than the overall results in Table~\ref{tab:class_balanced_metrics}, suggesting that knowledge retrieval and interpretation tasks are more challenging for most evaluated models.

\begin{table}[!htbp]
\centering
\caption{Evaluation Metrics for Knowledge Retrieval and Interpretation Tasks}
\label{tab:knowledge_class_balanced_metrics}
\small
{
\setlength{\tabcolsep}{5pt}
\renewcommand{\arraystretch}{1.15}
\makebox[\textwidth][c]{%
\begin{tabular}{l c c c c @{\hspace{3pt}} c @{\hspace{3pt}} c @{\hspace{3pt}} c}
\toprule
\textbf{Model}
& \textbf{Accuracy}
& \textbf{Approach}
& \begin{tabular}{c}\textbf{Source}\\\textbf{Validity}\end{tabular}
& \textbf{Tokens}
& \begin{tabular}{c}\textbf{Tool}\\\textbf{Calls}\end{tabular}
& \begin{tabular}{c}\textbf{Cost}\\\textbf{Est. (\$)}\end{tabular}
& \begin{tabular}{c}\textbf{Failure}\\\textbf{Rate (\%)}\end{tabular} \\
\midrule
Claude Sonnet 4.6
& $0.42 \pm 0.07$ & $3.65 \pm 0.22$ & $3.64 \pm 0.37$
& 198k & 4.2 & 0.62 & 0 \\

Qwen3-Max-Thinking
& $0.31 \pm 0.05$ & $3.50 \pm 0.23$ & $2.43 \pm 0.18$
& 181k & 3.4 & 0.22 & 1.2 \\

DeepSeek V3.2
& $0.33 \pm 0.06$ & $2.98 \pm 0.15$ & $2.39 \pm 0.27$
& 353k & 6.9 & 0.06 & 9.3 \\

Kimi-K2.5
& $0.36 \pm 0.05$ & $3.51 \pm 0.22$ & $2.82 \pm 0.25$
& 234k & 5.1 & 0.04 & 2.3 \\

Gemini-3.1-Pro
& $0.44 \pm 0.15$ & $3.67 \pm 0.22$ & $2.89 \pm 0.65$
& 245k & 3.5 & 0.16 & 2.3 \\

GPT-5-mini
& $0.23 \pm 0.06$ & $3.22 \pm 0.25$ & $3.15 \pm 0.72$
& 102k & 2.1 & 0.02 & 10.5 \\

GPT-5.2
& $0.45 \pm 0.11$ & $3.53 \pm 0.34$ & $4.25 \pm 0.65$
& 143k & 4.3 & 0.12 & 1.2 \\
\bottomrule
\end{tabular}%
}
}
\end{table}

\subsection{Results for Advanced Quantitative Modeling and Decision Analytics Tasks}
\label{sec:quant}

Table A3 also shows lower accuracy scores and significantly higher failure rates, showing that the questions under this category are more challenging compared to the other two categories. However, Gemini-3.1-Pro appears to perform better in his category and the comes with high token usage.

\begin{table}[!htbp]
\centering
\caption{Evaluation Metrics for Advanced Quantitative Modeling and Decision Analytics Tasks}
\label{tab:quant_class_balanced_metrics}
\small
{
\setlength{\tabcolsep}{5pt}
\renewcommand{\arraystretch}{1.15}
\makebox[\textwidth][c]{%
\begin{tabular}{l c c c c @{\hspace{3pt}} c @{\hspace{3pt}} c @{\hspace{3pt}} c}
\toprule
\textbf{Model}
& \textbf{Accuracy}
& \textbf{Approach}
& \begin{tabular}{c}\textbf{Source}\\\textbf{Validity}\end{tabular}
& \textbf{Tokens}
& \begin{tabular}{c}\textbf{Tool}\\\textbf{Calls}\end{tabular}
& \begin{tabular}{c}\textbf{Cost}\\\textbf{Est. (\$)}\end{tabular}
& \begin{tabular}{c}\textbf{Failure}\\\textbf{Rate (\%)}\end{tabular} \\
\midrule
Claude Sonnet 4.6
& $0.53 \pm 0.14$ & $4.32 \pm 0.21$ & $1.76 \pm 0.48$
& 391k & 14 & 1.28 & 8 \\

Qwen3-Max-Thinking
& $0.28 \pm 0.15$ & $3.42 \pm 0.45$ & $1.72 \pm 0.41$
& 474k & 13.9 & 0.60 & 6 \\

DeepSeek V3.2
& $0.29 \pm 0.18$ & $3.30 \pm 0.32$ & $1.86 \pm 0.55$
& 580k & 21 & 0.09 & 58 \\

Kimi-K2.5
& $0.44 \pm 0.14$ & $3.60 \pm 0.27$ & $2.21 \pm 0.42$
& 743k & 21 & 0.12 & 50 \\

Gemini-3.1-Pro
& $0.66 \pm 0.13$ & $3.39 \pm 0.47$ & $2.48 \pm 0.51$
& 783k & 12.4 & 0.48 & 8 \\

GPT-5-mini
& $0.37 \pm 0.17$ & $2.92 \pm 0.39$ & $3.21 \pm 0.55$
& 87k & 4.7 & 0.01 & 6 \\

GPT-5.2
& $0.49 \pm 0.16$ & $3.35 \pm 0.32$ & $3.36 \pm 0.49$
& 121k & 7.2 & 0.10 & 2 \\
\bottomrule
\end{tabular}%
}
}
\end{table}

\subsection{Judge Attribution Sanity Check for Gemini-3.1-Pro Accuracy Wins}
\label{sec:gemini_judge_attribution}

To assess whether Gemini-3.1-Pro's accuracy performance was disproportionately influenced by any single judge, we examined the subset of questions where Gemini-3.1-Pro achieved a median accuracy score greater than or equal to every other evaluated model. This subset includes 139 of 243 questions, corresponding to 57.2\% of the benchmark. Ties at the top are included.

\begin{table}[!htbp]
\centering
\caption{Judge Attribution for Gemini 3.1 Pro Accuracy Wins}
\label{tab:gemini_judge_attribution}

{
\setlength{\tabcolsep}{5pt}
\renewcommand{\arraystretch}{1.08}

\begin{tabular}{lcc}
\toprule
\textbf{Judge}
& \begin{tabular}{c}\textbf{Subset}\\\textbf{Gemini Wins, $n=139$}\end{tabular}
& \begin{tabular}{c}\textbf{Baseline}\\\textbf{All Gemini Panels, $n=243$}\end{tabular} \\
\midrule
OpenAI GPT-5-mini              & 32.01\% & 30.38\% \\
DeepInfra DeepSeek V3.2        & 28.06\% & 28.53\% \\
Google Gemini 3.1 Flash Lite   & 39.93\% & 41.08\% \\
\midrule
Total                          & 100.00\% & 100.00\% \\
\bottomrule
\end{tabular}
}
\end{table}

\begin{table}[!htbp]
\centering
\caption{Panel Tie Regimes for Gemini 3.1 Pro Accuracy Wins}
\label{tab:gemini_tie_regime}

{
\setlength{\tabcolsep}{6pt}
\renewcommand{\arraystretch}{1.08}

\begin{tabular}{lcc}
\toprule
\textbf{Regime}
& \textbf{Panels}
& \textbf{Share} \\
\midrule
3-way tie among judges       & 69 & 49.64\% \\
2-way tie with one outlier   & 51 & 36.69\% \\
All three judges different   & 19 & 13.67\% \\
\bottomrule
\end{tabular}
}
\end{table}

The attribution pattern does not suggest that Gemini 3.1 Pro's accuracy wins are driven by a single judge. The judge shares in the Gemini-winning subset are close to the baseline shares over all Gemini panels: OpenAI GPT-5-mini accounts for 32.01\% of the winning subset versus 30.38\% overall, DeepSeek V3.2 accounts for 28.06\% versus 28.53\%, and Gemini 3.1 Flash Lite accounts for 39.93\% versus 41.08\%. These differences are small, with all deviations within approximately two percentage points.

The tie-regime analysis provides an additional check. In 49.64\% of Gemini's winning panels, all three judges assigned the same median-relevant accuracy score. A further 36.69\% involved a two-judge agreement with one outlier. Thus, in 86.33\% of the Gemini-winning panels, at least two judges agreed on the score. Only 13.67\% of the winning panels had all three judges disagree, which are the cases where the median is most sensitive to a single judge. Overall, this suggests that Gemini 3.1 Pro's accuracy wins are primarily supported by cross-judge agreement rather than by judge-specific bias.

\subsection{Public framework and data repository overview}
\label{sec:repo_overview}
The public repository (https://github.com/Tume-AI/energy-evals) contains a complete list of the 212 questions. However, benchmark traces, ground truths, evaluation results and justifications are released for a subset containing 30 questions. This is to prevent data contamination and model overfitting issues. As subsequent versions of \textsc{EnergyEvals} become available, the datasets and codes will be updated. 

\subsection{Task IDs for source and tools impact analysis}
\label{sec:source_task_ids}
The task IDs for the 61 tasks without sources considered for the source impact analysis are as follows. The ids follow the format: \textit{With Source (Without Source Counterpart)}.

\begin{tcolorbox}[promptbox, title = With and Without source task IDs]
\ttfamily
\small
1 (22), 2 (23), 3 (24), 6 (27), 15 (36), 16 (37), 17 (38), 18 (39), 19 (40), 20 (41), 21 (42), 67 (45), 68 (46), 69 (47), 70 (48), 71 (49), 72 (50), 73 (51), 83 (61), 84 (62), 85 (63), 86 (64), 87 (101), 88 (102), 89 (103), 90 (104), 91 (105), 92 (106), 93 (107), 95 (109), 96 (110), 97 (111), 98 (112), 100 (114), 123 (152), 124 (153), 125 (154), 126 (155), 127 (156), 128 (157), 129 (158), 130 (159), 131 (160), 132 (161), 133 (162), 134 (163), 144 (172), 149 (177), 150 (178), 151 (179), 188 (213), 189 (214), 191 (216), 196 (221), 197 (222), 198 (223), 200 (225), 201 (226), 202 (227), 203 (228), 206 (231)
\end{tcolorbox}




The task IDs for the 30 tasks considered for the tool impact analysis are as follows.

\begin{tcolorbox}[promptbox, title = Tool impact analysis task IDs]
\ttfamily
\small
111, 112, 115, 172, 207, 209, 213, 214, 215, 216, 217, 218, 219, 220, 221, 222, 223, 224, 225, 226, 227, 228, 229, 230, 231, 232, 233, 234, 235, 245
\end{tcolorbox}

\clearpage

\subsection{Sample trace path of a pair of questions with and without sources}
The trace paths for the answers to questions 88 (with source) and 102 (without source variant) across all seven models evaluated are shown in the figure below.
\label{sec:trace_path}
\begin{figure}[h]
\centering
\includegraphics[width=1\linewidth]{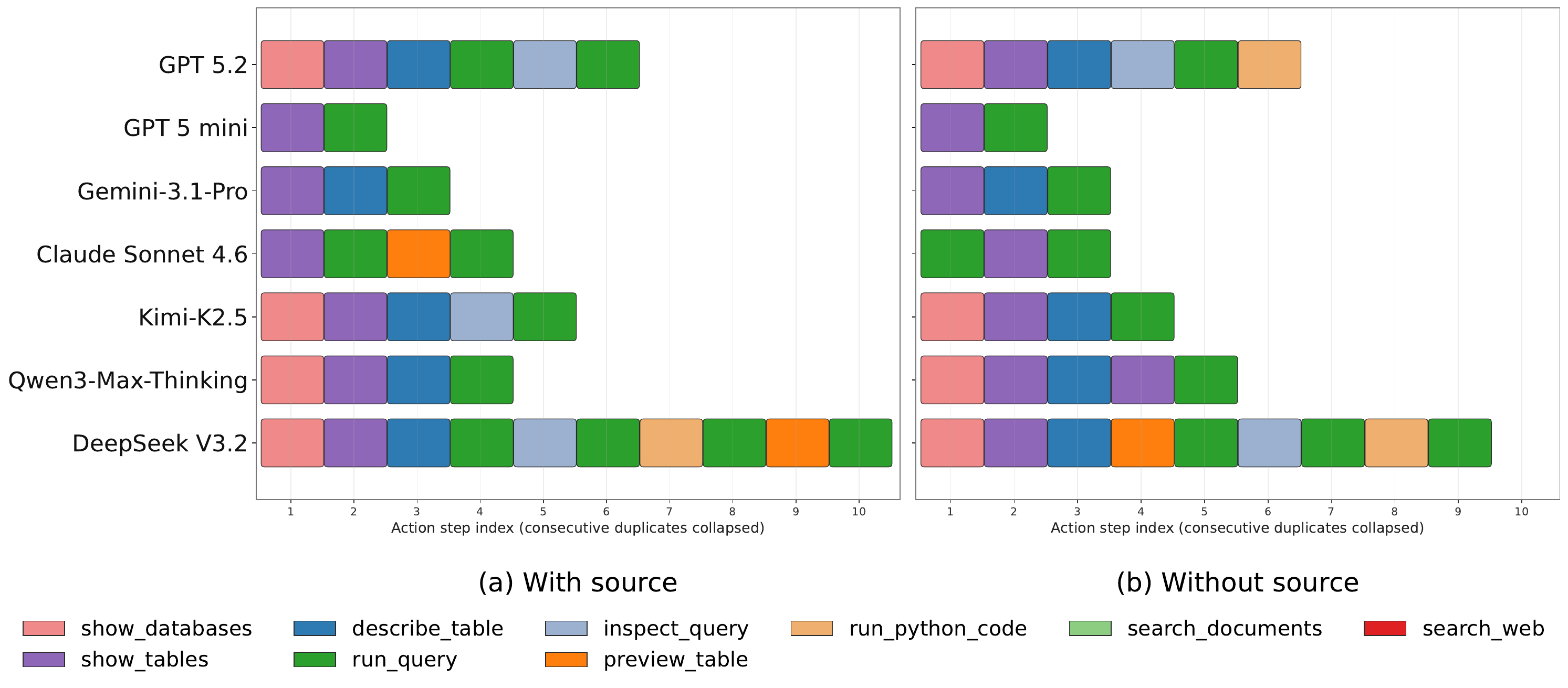}
\caption{Trace paths for all 7 models to answer the question (a) \textit{"What was the difference between average ERCOT weekday and weekend day-ahead prices in the summer of 2023 based on your ERCOT database?"} and (b) \textit{"What was the difference between average ERCOT weekday and weekend day-ahead prices in the summer of 2023?}}
\label{fig:trace_path}
\end{figure}

\end{document}